\documentclass[letterpaper]{article} 
\usepackage{aaai23}  
\usepackage{times}  
\usepackage{helvet}  
\usepackage{courier}  
\usepackage[hyphens]{url}  
\usepackage{graphicx} 
\usepackage{natbib}  
\usepackage{caption} 
\usepackage{algorithm}
\usepackage{algorithmic}
\usepackage{multirow}
\nocopyright
\usepackage{newfloat}
\usepackage{listings}
\usepackage{amsmath}
\usepackage{amssymb}
\usepackage{leftidx}
\usepackage{subfigure}
\usepackage{booktabs}
\usepackage[table]{xcolor}

\DeclareMathAlphabet{\mathbbold}{U}{bbold}{m}{n}

\DeclareCaptionStyle{ruled}{labelfont=normalfont,labelsep=colon,strut=off} 
\floatstyle{ruled}
\newfloat{listing}{tb}{lst}{}
\floatname{listing}{Listing}
\title{Deep Active Learning with Contrastive Learning Under Realistic Data Pool Assumptions}
\author{
    Jihyo Kim,\textsuperscript{\rm 1}
    Jeonghyeon Kim,\textsuperscript{\rm 2}
    Sangheum Hwang\textsuperscript{\rm 2}\protect{\thanks{Corresponding author}}
}
\affiliations{
    \textsuperscript{\rm 1} Department of Data Science, Seoul National University of Science and Technology\\

    \textsuperscript{\rm 2} Department of Industrial Engineering, Seoul National University of Science and Technology\\
    jihyo.kim@ds.seoultech.ac.kr, mawjdgus@seoultech.ac.kr, shwang@seoultech.ac.kr

}

\begin{document}

\maketitle

\begin{abstract}
Active learning aims to identify the most informative data from an unlabeled data pool that enables a model to reach the desired accuracy rapidly. This benefits especially deep neural networks which generally require a huge number of labeled samples to achieve high performance. Most existing active learning methods have been evaluated in an ideal setting where only samples relevant to the target task, i.e., in-distribution samples, exist in an unlabeled data pool. A data pool gathered from the wild, however, is likely to include samples that are irrelevant to the target task at all and/or too ambiguous to assign a single class label even for the oracle. We argue that assuming an unlabeled data pool consisting of samples from various distributions is more realistic. In this work, we introduce new active learning benchmarks that include ambiguous, task-irrelevant out-of-distribution as well as in-distribution samples. We also propose an active learning method designed to acquire informative in-distribution samples in priority. The proposed method leverages both labeled and unlabeled data pools and selects samples from clusters on the feature space constructed via contrastive learning. Experimental results demonstrate that the proposed method requires a lower annotation budget than existing active learning methods to reach the same level of accuracy.
\end{abstract}

\begin{figure}[!t]
\centerline{\includegraphics[width=\columnwidth]{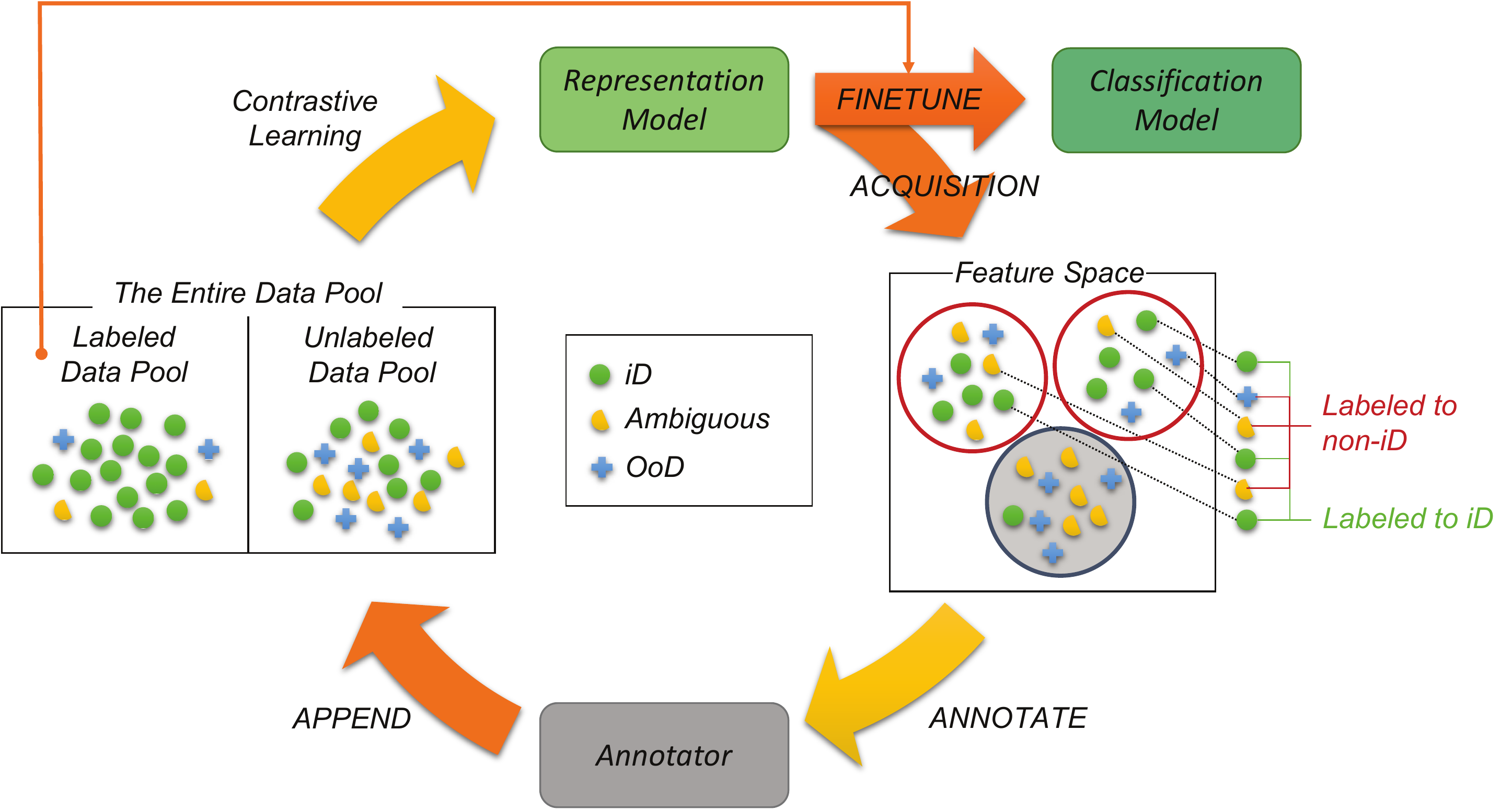}}
\caption{The framework of the proposed active learning method. A representation model is trained with both labeled and unlabeled data pools by contrastive learning. The sample acquisition is processed on the contrastively learned feature space. The selected samples are labeled by the annotator, and the labeled data pool is updated by adding the newly labeled samples including non-iD. Then, the representation model is finetuned with the labeled data pool. The finetuned representation model performs as a classification model that makes predictions for the given task.}
\label{fig:framework}
\end{figure}

\section{Introduction}
Deep neural networks have achieved numerous successes across a wide range of tasks. However, data labeling is still a major obstacle to utilizing deep learning models in practice, especially in specialized domains such as medicine where labeling costs are expensive, since they usually require a lot of labeled samples to achieve good performance.

Active learning is a machine learning framework designed to alleviate the labeling cost issue~\cite{settles2009active}. In the active learning process, a deep neural network chooses a limited number of samples that are expected to be the most beneficial for improving task performance. An annotator (i.e., oracle) labels the chosen samples, and then those labeled samples are used for updating the current model. The criteria of most existing active learning methods for estimating the benefit of samples are designed to consider informativeness, representativeness, or both~\cite{sinha2019variational,sener2018active,kirsch2019batchbald}.

Most of the existing active learning methods have been evaluated on an unlabeled pool constructed using a single benchmark dataset. For example, an unlabeled data pool can be a subset of MNIST training dataset. This unlabeled data pool is clearly made up of samples from a single identical distribution. However, we cannot ensure that an unlabeled data pool collected in the wild has only samples from an identical distribution. In practice, an unlabeled data pool is more likely to contain samples that are not worth labeling. It is reasonable to assume that such an unlabeled pool might consist of roughly three categories: \romannumeral 1 ) task-relevant samples with a single clear label (in-distribution; iD), which potentially contribute to improving the target task performance, \romannumeral 2 ) samples that are too ambiguous to be labeled as one of the target classes, and \romannumeral 3 ) task-irrelevant samples containing far different semantics with iD (out-of-distribution; OoD). In the active learning process, it is desirable to acquire informative samples from the iD category.

There is a challenge in employing existing active learning methods directly to such an unlabeled data pool. For example, active learning methods generally assign low confidence values to samples from unseen distributions, therefore OoD samples are highly likely to be selected with the acquisition criterion such as least confidence or entropy. Furthermore, ambiguous samples are easily confused with hard-to-classify iD samples which are the most valuable ones to boost performance. As a result, the acquired batch to be labeled will be dominated by non-iD samples although iD samples, especially hard-to-classify ones, should be acquired. If this batch, which consists mostly of non-iD samples, is delivered to the annotator, a given annotation budget is wasted since most of the cost will be spent on filtering out non-iD samples. In other words, this batch just consumes the annotation budget without any contributions to the goal of active learning.

To consider the challenging yet realistic unlabeled pool, we propose new active learning benchmarks based on an unlabeled data pool consisting of samples from the aforementioned three categories, iD, ambiguous, and OoD samples. Especially, we define ambiguous samples as samples for which it is difficult to allocate a clear single label (e.g., poorly written digits $1$ vs. $7$, Siberian husky vs. Alaskan malamute). This ambiguity makes it more difficult for deep neural networks to choose informative iD samples. Concretely, we construct two benchmarks, MixMNIST and MixCIFAR60, considering the task difficulty level.

We also present an active learning method designed to perform properly on the realistic active learning benchmarks. To distinguish iD samples from non-iD samples and select informative iD ones, the proposed method leverages both labeled and unlabeled samples by using the contrastive learning approach. Our sample acquisition strategy utilizes clusters obtained by $k$-means clustering in the contrastively learned feature space. We choose samples based on two assumptions: \romannumeral 1 ) in clusters where non-iD samples are not the majority, iD samples are closer to the centroid than non-iD samples, \romannumeral 2 ) samples far from the centroid provide informative knowledge that a model has not yet learned. Based on these assumptions, for each cluster, we select unlabeled samples that are sufficiently far from the centroid, and at the same time, are closer to the centroid than samples labeled as non-iD.

We evaluate our active learning method on the proposed benchmarks. For precise performance comparison, we consider the total annotation costs, which include not only labeling costs for iD samples but wasted costs to filter out non-iD samples. The experimental results reveal that the proposed method spends significantly lower annotation costs than the existing methods while showing comparable performance on the target task.

Our contributions are summarized as:
\begin{itemize}
    \item We address the necessity of consideration of an unlabeled data pool containing samples from diverse distributions. The unlabeled data pool with this configuration can significantly increase the annotation costs of existing active learning methods.
    \item We propose new realistic active learning benchmarks, named MixMNIST and MixCIFAR60, consisting of samples from three categories: in-distribution, ambiguous, and out-of-distribution samples.
    \item We propose an active learning method for acquiring informative iD samples based on clusters over features extracted from a model trained via contrastive learning. The proposed method consumes considerably low annotation costs while showing comparable accuracy.
\end{itemize}

\begin{table}
    \centering
    \caption{The number of samples belonging to each category in our benchmark datasets.}
    \label{tab:dataset-stat}
    \begin{tabular}{c|c|c|c|c} 
    \toprule
    & {\footnotesize{\textbf{iD}}} & \footnotesize{\textbf{Ambiguous}} & \multicolumn{2}{c}{\footnotesize{\textbf{OoD}}} 
    \\\cmidrule{1-5}
    \multirow{4}{*}{\rotatebox[origin=c]{90}{\scriptsize{\textbf{MixMNIST}}}}
     & \multirow{2}{*}{\footnotesize{MNIST}} & \multirow{2}{*}{\footnotesize{NVAE-MNIST}} & \multicolumn{2}{c}{\multirow{2}{*}{\footnotesize{notMNIST}}} \\&&&\multicolumn{2}{c}{}\\ \cmidrule{2-5}
    & \multirow{2}{*}{60,000} & \multirow{2}{*}{15,000} & \multicolumn{2}{c}{\multirow{2}{*}{18,724}} \\&&&\multicolumn{2}{c}{}\\ \cmidrule{1-5}
    \multirow{4}{*}{\rotatebox[origin=c]{90}{\scriptsize{\textbf{MixCIFAR60}}}} & \multirow{2}{*}{\footnotesize{CIFAR60}} & \multirow{2}{*}{\footnotesize{CIFAR40}} & \multirow{2}{*}{\footnotesize{LSUN-FIX}} & \multirow{2}{*}{\footnotesize{SVHN}} \\ &&&\\ \cmidrule{2-5}
    & \multirow{2}{*}{30,000} & \multirow{2}{*}{15,000} & \multirow{2}{*}{7,500} & \multirow{2}{*}{7,500} \\&&&\\
    \bottomrule
    \end{tabular}
\end{table}

\section{Related Work}
In this section, we briefly review the related works in the field of active learning and contrastive learning.

\subsubsection{Active learning.} Common sampling strategies in active learning are uncertainty-based and representative-based methods. Uncertainty-based sampling methods select a sample based on predictive uncertainty measures such as least confidence and entropy~\cite{settles2009active}. Representative-based sampling methods aim at finding diverse samples that approximate the true data distribution well, e.g., it can be achieved by solving the set covering problem~\cite{sener2018active}.

The majority of previous works on active learning have proposed effective sampling methods. However, they assume that an ideal unlabeled data pool consisting of only iD samples is given, which is hardly expected in real-world scenarios. Recently, several studies have attempted to set up realistic experimental settings, taking into account imbalances, redundancies, and open-set~\cite{mukhoti2021deep, mandivarapu2022deep, kothawade2021similar, ning2022active}. Deep Deterministic Uncertainty (DDU)~\cite{mukhoti2021deep} constructs the unlabeled data pool containing ambiguous handwritten digits, AmbiguousMNIST, to demonstrate the effectiveness of their proposed method that disentangles the total uncertainty into aleatoric and epistemic ones.
In this work, we consider more challenging settings where both ambiguous and OoD samples exist in a given unlabeled data pool.

\subsubsection{Contrastive learning.} Contrastive learning can be applied to both self-supervised and supervised learning problems~\cite{chen2020simple,khosla2020supervised}. In the self-supervised setting, SimCLR~\cite{chen2020simple} learns representations by maximizing agreement between different augmentations of the same data point via contrastive loss. 
Another class of contrastive learning is supervised contrastive learning which is structurally similar to the self-supervised approach, but features of the same class are attracted closer together than those of different classes~\cite{khosla2020supervised}. Therefore, it can be seen as an extension of self-supervised contrastive learning, allowing the exploitation of label information.

In our proposed benchmark settings, a model can encounter both labeled and unlabeled samples from various distributions. 
To learn feature representations that have the capability of distinguishing non-iD samples from iD ones, we utilize self-supervised and supervised contrastive losses to unlabeled and labeled samples, respectively.

\section{Unlabeled Data Pool in the Wild}
Our proposed benchmarks are built upon the simplified real-world environment, i.e., unlabeled samples belong to one of the aforementioned categories, iD, ambiguous, or OoD. For simplicity, we set the ratios of iD, ambiguous and OoD samples in each benchmark to about $4:1:1$. We create two benchmark datasets, named MixMNIST and MixCIFAR60, based on commonly used MNIST and CIFAR100 in active learning. Table \ref{tab:dataset-stat} summarizes the number of samples belonging to each category in our benchmark datasets. The following sections provide a detailed description of how we construct new benchmark datasets.

\subsection{MixMNIST}
We set MNIST as the iD dataset for MixMNIST, so our target task is to classify a given image into 10 digit classes. All $60,000$ training images are included in the unlabeled data pool as the iD samples, and $10,000$ test images are used to evaluate the model performance. The OoD samples are from notMNIST dataset~\footnote{\url{http://yaroslavvb.blogspot.com/2011/09/notmnist-dataset.html}}. They are distinguishable from MNIST because the images of notMNIST correspond to letters from A to J. We use $18,724$ notMNIST images. 

\begin{figure}[!ht]
    \minipage{0.33\columnwidth}
      \includegraphics[width=0.99\columnwidth]{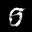}
      \label{fig:nvae-mnist-1}
    \endminipage
    \minipage{0.33\columnwidth}
      \includegraphics[width=0.99\columnwidth]{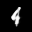}
      \label{fig:nvae-mnist-2}
    \endminipage
    \minipage{0.33\columnwidth}%
      \includegraphics[width=0.99\columnwidth]{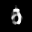}
      \label{fig:nvae-mnist-3}
    \endminipage
\caption{NVAE-MNIST image examples generated via NVAE. Two randomly selected latent vectors from the encoder are linearly interpolated, and the pretrained NVAE decoder generates ambiguous images with those vectors.}
\label{fig:nvae-mnist}
\end{figure}

To give ambiguity to images, we generate MNIST-like images using Nouveau VAE (NVAE)~\cite{vahdat2020nvae}. We adopt MNIST pretrained models from their github repository~\footnote{\url{https://github.com/NVlabs/NVAE}}. Two randomly selected latent vectors from the encoder are linearly interpolated in a $7:3$ ratio, and then fed to the decoder. Since we interpolate two latent vectors from the encoder, the generated samples have some degree of ambiguity in labeling. We examine the generated images with ten different MNIST classifiers to evaluate the ambiguity level. Specifically, the generated image is filtered if all ten models predict the same class. These images are thought to be clear of any ambiguity. Images with four or more diverse predictions are also filtered. We consider these images to be inappropriately generated images. Figure~\ref{fig:nvae-mnist} shows examples of our NVAE-MNIST images. For the left image, seven out of the ten MNIST classifiers predict the class $0$, two predict the class $6$, and one predicts the class $8$. Following this procedure, we construct NVAE-MNIST consisting of $15,000$ ambiguous MNIST-like images.

\subsection{MixCIFAR60}
MixCIFAR60 is based on CIFAR100 which is another popular dataset in active learning. The 100 classes of CIFAR100 can be grouped into 20 superclasses with each having five subclasses. \citet{kim2021unified} designed the iD and near-OoD datasets by splitting each superclass into two groups, so there might be shared semantics between these two groups although they have no overlapping classes in a fine-grained level. These shared semantics make it difficult for the oracle to assign a single class label with high certainty. Hence, we consider the near-OoD dataset as ambiguous images.

Resized LSUN with 10 scene classes is a popular OoD dataset against CIFAR100 in OoD detection (OoDD) task. However, as \citet{tack2020csi} pointed out, images in resized LSUN contain artificial noises caused by an improper resizing process. These noises can be used as shortcuts which result in overestimated detection performance. Therefore, we use the reconstructed LSUN, referred to as LSUN-FIX, following~\citet{tack2020csi}. Without consideration of class information, we randomly sampled $7,500$ images from LSUN-FIX. SVHN, also another popular OoD dataset, consists of 10 digit classes taken from house-number color images. We uniformly select $7,500$ images at random. The total $15,000$ images from LSUN-FIX and SVHN are used as OoD samples in MixCIFAR60.

\section{Active Learning via Contrastive Learning}
In this work, we employ a contrastive learning method to fully leverage samples in both labeled and unlabeled data pools. Most active learning literature employs a supervised training approach. In a supervised training approach, a model is trained with only samples in the labeled data pool even if numerous samples in the unlabeled data pool are accessible. With samples in the unlabeled data pool, our model tries to discriminate every instance via contrastive loss, and thereby, the instances sharing high-level semantics can be clustered on the learned feature space. This is highly desirable, especially for the unlabeled data pool consisting of various sources (i.e., iD, ambiguous, and OoD) since one can effectively identify non-iD samples based on distances among features. Contrastive learning also benefits from samples in the labeled data pool. The instance with the same label will be pulled together in the feature space, resulting in dense clusters. By leveraging this property, we can expect that iD and non-iD samples are more easily distinguishable in the feature space. We additionally consider an auxiliary class for non-iD samples. This aids in attracting non-iD samples to a single cluster so that the feature space can be learned in a way to clearly separate iD and non-iD samples. Figure~\ref{fig:framework} depicts the overall framework of our proposed method.

The following sections describe how to train a representation model with both data pools via contrastive learning, and the acquisition strategy based on the learned feature representations.

\subsection{Contrastive learning with both data pools}
To train a representation model with both a labeled data pool $\mathcal{P}^\mathcal{L}$ and an unlabeled data pool $\mathcal{P}^\mathcal{U}$, we consider two contrastive learning methods simultaneously: self-supervised contrastive learning~\cite{chen2020simple} for $\mathcal{P}^\mathcal{U}$ and supervised contrastive (SupCon) learning~\cite{khosla2020supervised} for $\mathcal{P}^\mathcal{L}$. We employ contrastive loss which is computed as:
\begin{align}
\label{formula: con-loss}
    && \mathcal{L}(\mathbf{z}_i,\mathbf{z}_j) = - \log \frac{
    \exp(\langle\mathbf{z}_i,\mathbf{z}_j\rangle/\tau)}
    {\sum_{k \in 2N \backslash i} {\exp(\langle\mathbf{z}_i, \mathbf{z}_k \rangle /\tau)}}
\end{align}
where $\langle \cdot , \cdot \rangle$ is a similarity function, $N$ is a mini-batch size, and $\tau$ means a temperature.
For unlabeled samples in $\mathcal{P}^\mathcal{U}$, $\textbf{z}_i$ and $\textbf{z}_j$ denote feature vectors of different augmented views from the same input data after a projection head. In contrast, for labeled samples in $\mathcal{P}^\mathcal{L}$, $\textbf{z}_i$ and $\textbf{z}_j$ denote feature vectors of different input data from the same class. For the clarity, we denote contrastive loss for $\mathcal{P}^\mathcal{U}$ as $\mathcal{L}_\text{con}$ and that for $\mathcal{P}^\mathcal{L}$ as $\mathcal{L}_\text{supcon}$. Then, the total loss $\mathcal{L}_\text{total}$ is defined as:
\begin{align}
\label{formula:total-loss}
    \mathcal{L}_\text{total} = \mathcal{L}_\text{con} + \mathcal{L}_\text{supcon}
\end{align}

\begin{figure}[!t]
    \centerline{\includegraphics[width=\columnwidth]{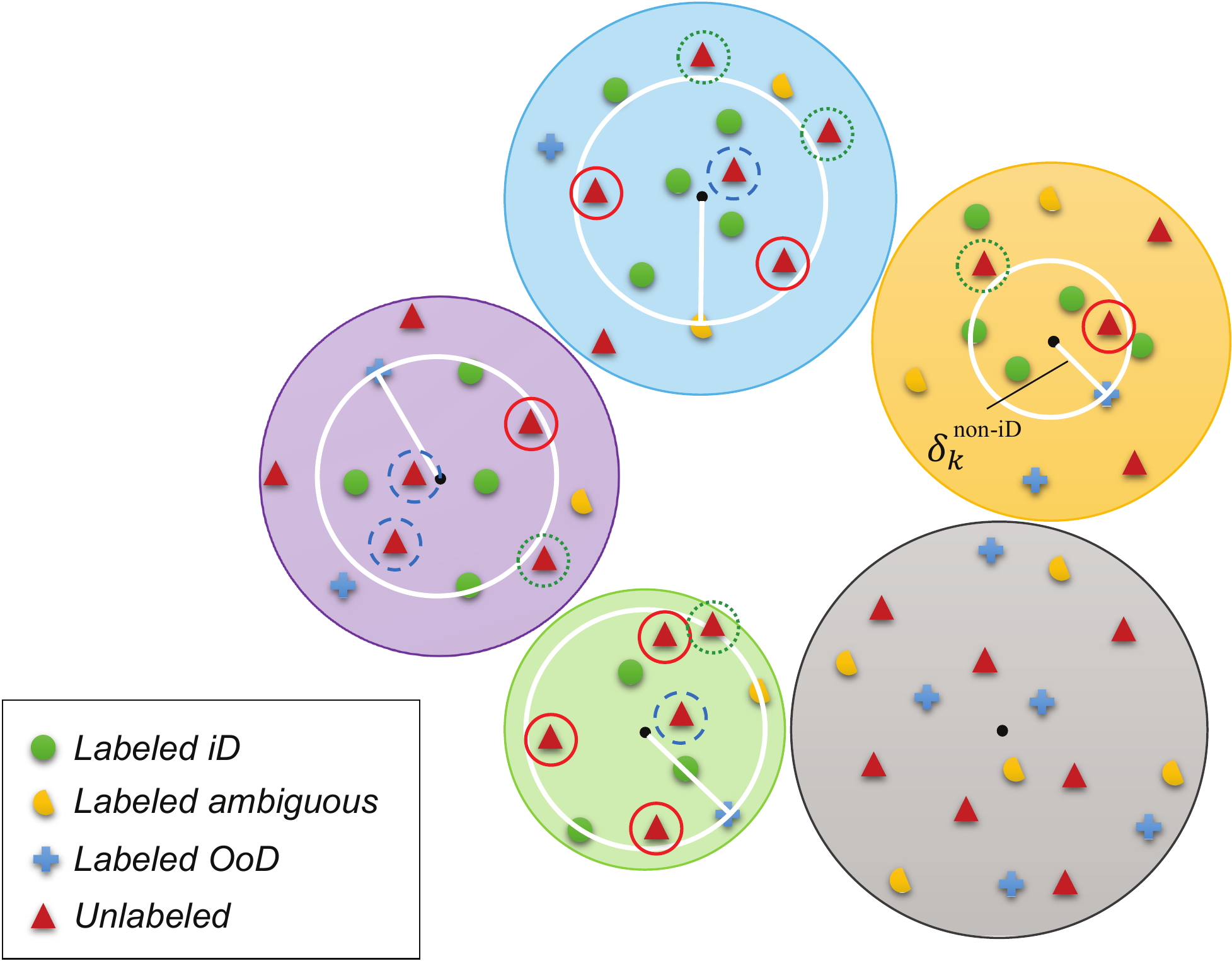}}
    \caption{The concept of the proposed acquisition strategy. The samples are acquired using feature clusters.}
    \label{fig:distance-clustering}
\end{figure}

\subsection{Acquisition strategy on the feature space}
The features of the trained representation model via contrastive learning contain valuable semantic knowledge of both $\mathcal{P}^\mathcal{U}$ and $\mathcal{P}^\mathcal{L}$. Leveraging this knowledge enables us to acquire informative samples on the learned feature space. We determine the samples to be acquired by a location within the cluster established on this feature space. Specifically, we select samples nearby the centroid-closest non-iD sample labeled by the oracle based on the aforementioned two assumptions in the Introduction section. 

Figure~\ref{fig:distance-clustering} describes the proposed acquisition strategy. 
Firstly, we examine all clusters to exclude a particular cluster that contains the largest number of labeled non-iD samples (i.e., the gray cluster) since it can be regarded as a cluster of non-iD samples. For the remaining clusters, we select samples from each cluster based on the distances between the cluster centroid and feature representations. The white line indicates the circle with the distance between the cluster centroid and the centroid-closest non-iD features $\delta_k^\text{non-iD}$ as its radius. The following samples are sequentially selected: \romannumeral 1 ) samples close to but inside the white circle (e.g., red solid circles), \romannumeral 2 ) samples inside the white circle regardless of proximity (e.g., blue dashed circles), and \romannumeral 3 ) samples close to but outside the white circle (e.g., green dotted circles).

Algorithm~\ref{alg:distance-clustering} presents the proposed acquisition process in detail. Let $\mathcal{P}^{\mathcal{L}}=\{\mathbf{x}^l_i, y^l_i\}_{i=1}^{m}$ be a labeled data pool consisting of $m$ samples, where $y^l_i \in \{1,2,3,...,K+1\}$ and $\mathcal{P}^{\mathcal{U}}=\{\mathbf{x}^u_j\}_{j=1}^{n}$ be an unlabeled data pool consisting of $n$ samples. We assign an auxiliary class $y^l_i=K+1$ if $x^l_i$ is an ambiguous or OoD sample. 
First, we build $K+1$ clusters via the $k$-means clustering using feature vectors $f(\mathbf{x};\mathbf{w})=\mathbf{h}$ for all $\mathbf{x} \in \mathcal{P}^{\mathcal{L}} \cup \mathcal{P}^{\mathcal{U}} $, where $f$ is a trained representation model with the total loss in Equation~\ref{formula:total-loss} and $\mathbf{w}$ is the representation model's parameters. Then we calculate the centroid $\mathbf{c}_k$ of each cluster $\mathcal{C}_k$ where $k \in \{1,2,3,...,K+1\}$. The cluster with the highest proportion of $\{\mathbf{x}^l_i\}$ whose ${y_i^l = K+1}$ is regarded as a cluster of non-iD samples and therefore, it is not taken into consideration for acquisition. Let $\delta_k(\mathbf{c}_k,\mathbf{h})$ be a distance between $\mathbf{c}_k$ and $\mathbf{h}$, and $\delta_k^{\text{non-iD}}$ be $\min_{\mathbf{h}_i^l \in \mathcal{C}_k} \delta_k(\mathbf{c}_k,\mathbf{h}_i^l)$ over all $\mathbf{h}_i^l$ of non-iD samples. The number of samples acquired from each cluster is determined by the ratio of $|\mathcal{C}_k|$. We repeatedly sample $\mathbf{x}_j^u \in \mathcal{C}_k$ whose $\delta_k(\mathbf{c}_k,\mathbf{h}^u_j) \leq \delta_k^\text{non-iD}$ (i.e., inside of the white circle in Figure~\ref{fig:distance-clustering}) if the difference $|\delta_k^\text{non-iD}-\delta_k(\mathbf{c}_k, \mathbf{h}_j^u)|$ is the minimum over all $\mathbf{h}_j^u$ until the number of iD samples annotated by the oracle meets the predefined sample size. If the number of the acquired iD samples still does not meets the predefined sample size, we sample $\mathbf{x}_j^u$ whose $\mathbf{h}_j^u$ is outside of the white circle (i.e., $\delta_k(\mathbf{c}_k,\mathbf{h}^u_j) > \delta_k^\text{non-iD}$) according to the sequence in which the difference $|\delta_k^\text{non-iD}-\delta_k(\mathbf{c}_k, \mathbf{h}_j^u)|$ is minimal.

\begin{algorithm}[tb]
\caption{The proposed acquisition function on stage $t$}
\label{alg:distance-clustering}
\textbf{Input}: Labeled pool $\mathcal{P^L}$,
Unlabeled pool $\mathcal{P^U}$, Cluster $\mathcal{C}_k$\\
\textbf{Output}: $\textit{Samples}_\text{iD} \cup \textit{Samples}_\text{non-iD}$
\begin{algorithmic}[1] 
\STATE $\mathbf{h}^l,\mathbf{h}^u\leftarrow$ Feature representation of samples in $\mathcal{P^L}$,$\mathcal{P^U}$
\STATE $\mathbf{c}_k\leftarrow$ Centroid of $\mathcal{C}_k$
\STATE $\mathbf{h}^n_k \leftarrow$ The non-iD sample in $\mathbf{h}^l$ closest to $\mathbf{c}_k$ of $\mathcal{C}_k$
\STATE $\delta_k(\mathbf{c}_k,\mathbf{h}^u)\leftarrow$ Distance between $\mathbf{h}^{u}$ and $\mathbf{c}_k$
\STATE $N\leftarrow$ The number of iD samples to be acquired
\STATE $N_k\leftarrow$ The number of samples to be acquired from $\mathcal{C}_k$
\STATE $\textit{Samples}_\text{iD},\textit{Samples}_\text{non-iD}$ $\leftarrow\varnothing,\varnothing$

\WHILE{$|\textit{Samples}_\text{iD}|<N$}
\FORALL{$k\in\{1,...,K+1\}$}
\STATE $S_k$ $\leftarrow\varnothing$, a set of samples acquired from $\mathcal{C}_k$
\STATE $S_\text{iD}$ $\leftarrow\varnothing$, a set of annotated iD samples
\STATE $S_\text{non-iD}$ $\leftarrow\varnothing$, a set of annotated non-iD samples
\STATE $\delta_k^\text{non-iD}\leftarrow$ $\delta_k(\mathbf{c}_k,\mathbf{h}^n_k)$
\WHILE{$|S_k|<N_k$}
\IF {exists a sample satisfying $\delta_k(\mathbf{c}_k,\mathbf{h}^u) \leq \delta_k^\text{non-iD}$}
\STATE Add maximum $\delta_k(\mathbf{c}_k,\mathbf{h}^u)$ sample to  $S_k$
\ELSE
\STATE Add minimum $\delta_k(\mathbf{c}_k,\mathbf{h}^u)$ sample to  $S_k$
\ENDIF
\ENDWHILE
\STATE \textit{Oracle} annotates samples in $S_k$
\STATE $S_\text{iD},S_\text{non-iD}\leftarrow$ Annotated iD and non-iD samples
\STATE $\textit{Samples}_\text{iD}\leftarrow \textit{Samples}_\text{iD}\cup S_\text{iD}$
\STATE $\textit{Samples}_\text{non-iD}\leftarrow \textit{Samples}_\text{non-iD}\cup S_\text{non-iD}$
\ENDFOR
\ENDWHILE
\STATE \textbf{return} $\textit{Samples}_\text{iD} \cup \textit{Samples}_\text{non-iD}$
\end{algorithmic}
\end{algorithm}

\section{Experimental Results}
We compare the proposed method to other commonly employed active learning methods with our MixMNIST and MixCIFAR60 benchmarks. The following subsequent sections present the experimental settings and the evaluation results.

\subsection{Experimental settings}

\subsubsection{Comparative methods.}
For MixMNIST, we examine three acquisition strategies, random, least confidence, and entropy sampling. Random sampling chooses samples uniformly at random from an unlabeled data pool. Least confidence sampling considers samples with low max-softmax values as the most informative samples. Entropy sampling selects samples associated with high predictive entropy as informative ones.
We use the three-convolutional-layer architecture. The model is trained during $100$ epochs using Adam with a learning rate of $1e^{-3}$, weight decay of $5e^{-3}$, and a batch size of $32$. The learning rate is reduced by a factor of $10$ at $70$ epochs and $90$ epochs. We regularize the model with gradient clipping with the norm of $2.5$ and label smoothing with the value of $0.1$ for training stability.

For MixCIFAR60, DDU~\cite{mukhoti2021deep} is additionally considered as a baseline along with the same three acquisition strategies (i.e., random, least confidence, and entropy sampling) in MixMNIST experiment. Consequently, the four comparative methods are evaluated with ResNet-18~\cite{he2016identity}. We train the model during $200$ epochs using SGD with a momentum of $0.9$, an initial learning rate of $0.1$, weight decay of $1e^{-4}$, and a batch size of $128$. The learning rate is reduced by a factor of $10$ at $120$ epochs and $160$ epochs. The standard data augmentation scheme is employed: random horizontal flip and $32\times32$ random crop after adding $4$ zero-pixels on each side of an image.

We consider one auxiliary class for non-iD samples for a fair comparison with our proposed method. The unlabeled non-iD samples which yield the maximum softmax value on the auxiliary class are filtered out from the acquisition process. This auxiliary class helps the model to considerably reduce the annotation costs by identifying non-iD samples. Therefore, introducing the auxiliary class makes the comparative methods strong baselines.

\subsubsection{The proposed method.}
We set random sampling from each cluster (Random(CL)) as the baseline of acquisition strategy on the contrastively learned feature space. The sampling size for each cluster is determined based on the cluster size ratio. The features for clustering are extracted from the last layer of the trained representation model.

At the very beginning of the active learning process (e.g., stage $0$) on MixMNIST, the representation model is trained with the entire unlabeled data pool during $300$ epochs using Adam with an initial learning rate of $3e^{-4}$, weight decay of $1e^{-4}$, the temperature of $0.07$ (refer to Equation.~\ref{formula: con-loss}) and a batch size of $1024$. The learning rate is reduced following cosine annealing used in \citet{chen2020simple}. After the stage $0$, the representation model is continuously trained with the updated both data pools during $100$ epochs without any changes to the hyperparameters. We use affine transformations such as rotation, shearing, and scaling for the data augmentation scheme of unlabeled images~\footnote{\url{https://github.com/mdiephuis/SimCLR}}.
To make predictions for the target task, we finetune the classification model which has an output linear layer followed by the representation model during $100$ epochs using Adam with a learning rate of $1e^{-3}$, weight decay of $1e^{-6}$, and a batch size of $32$. Label smoothing with value of $0.1$ is used to regularize the classification model. 

For MixCIFAR60, the representation model is trained with the entire unlabeled data pool during $500$ epochs and other hyperparameters are set to the same values in the MixMNIST experiment. After the stage $0$, we train the representation model continuously during $300$ epochs. The classification model is finetuned during longer epochs than MixMNIST, which is $200$, using SGD with a learning rate of $0.1$, no weight decay, and a batch size of $256$. The data augmentation schemes for unlabeled images are set following \citet{chen2020simple}, and  those for labeled images are the same as the settings for the comparative models.

\begin{figure}[!t]
\centering
    \subfigure[MixMNIST]{
        \includegraphics[width=0.85\columnwidth]{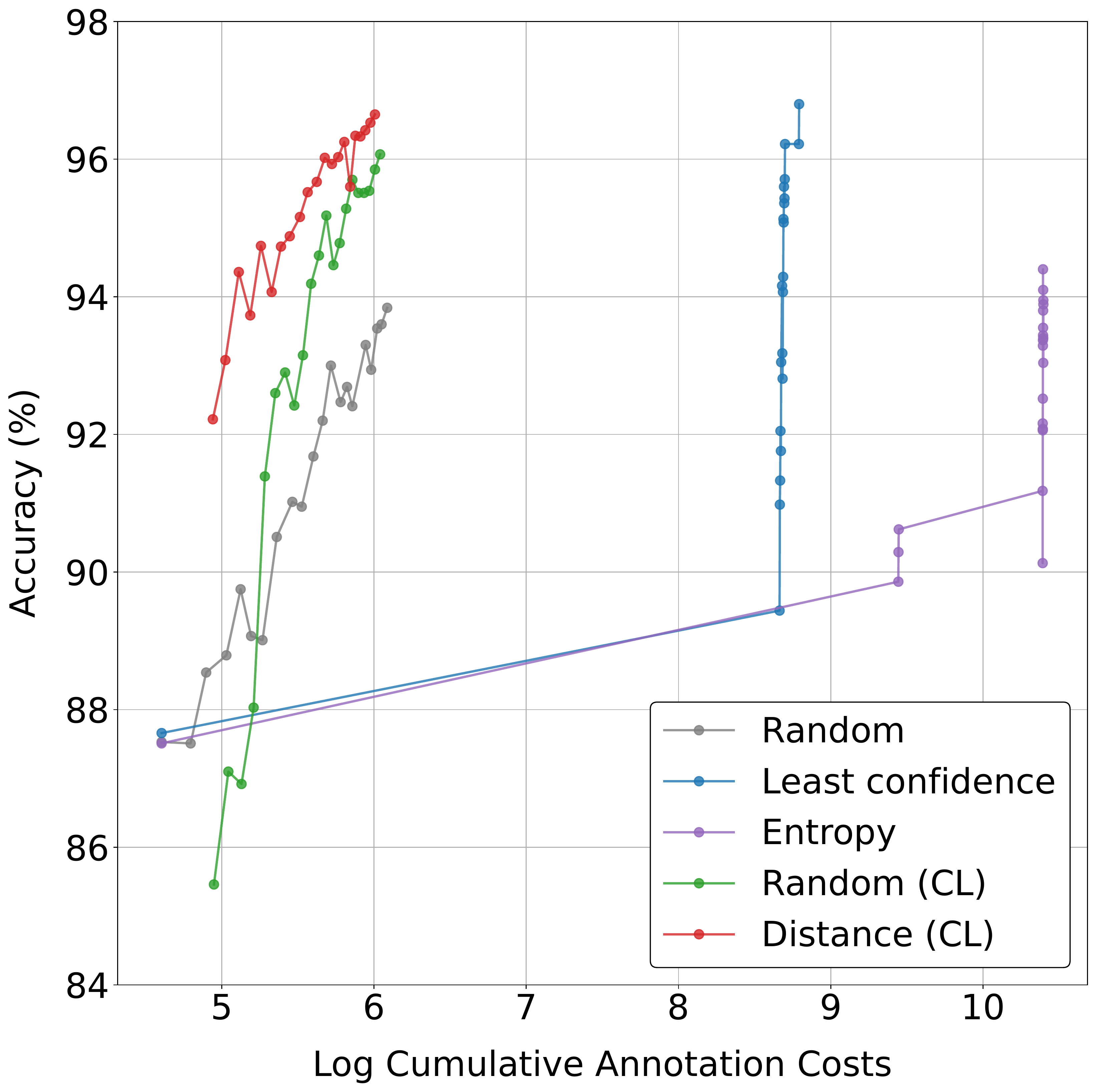}
        \label{fig:perf-mixedmnist}
    }
    \subfigure[MixCIFAR60]{
        \includegraphics[width=0.85\columnwidth]{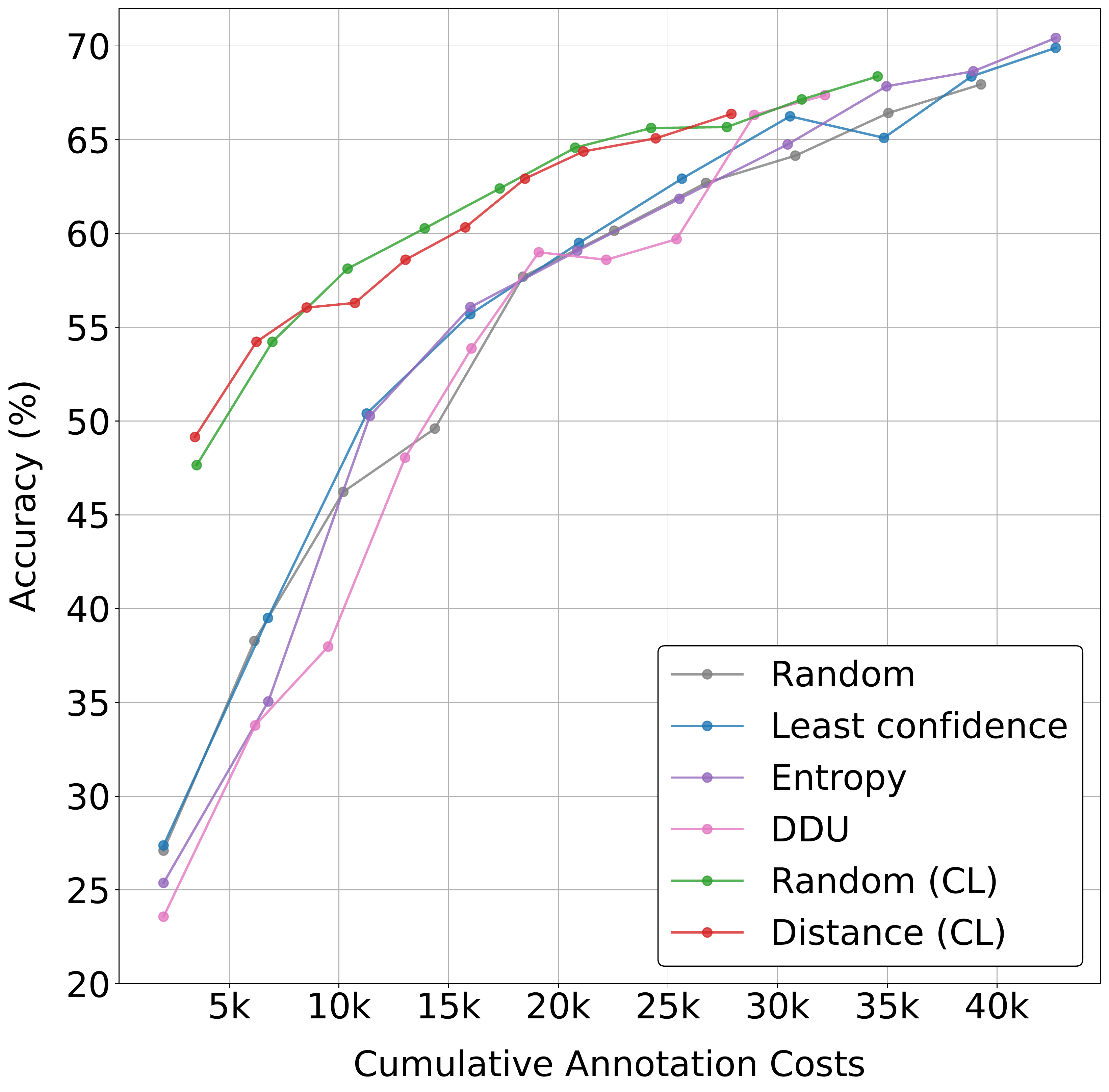}
        \label{fig:perf-mixedcifar60}
    }
    \caption{Active learning performance of the comparative methods. For MixMNIST, the annotation cost is log scaled.}
    \label{fig:performance}
\end{figure}

\subsubsection{Active learning.}
For MixMNIST, we have an initial unlabeled data pool $\mathcal{P}^\mathcal{U}_0$ which is the size of $93,724$, and an empty labeled iD data pool $\mathcal{P}^\mathcal{L}_0$ at the beginning of active learning process. We set the initial number of labeled iD samples to $100$. Before the start of the first stage, samples including $100$ iD samples and non-iD samples examined to acquire $100$ iD samples are removed from $\mathcal{P}^\mathcal{U}_0$, and updated to $\mathcal{P}^\mathcal{L}_1$. The minimum number of iD samples to be acquired at a single stage is set to $10$. At the end of every stage $t$, the $\mathcal{P}^\mathcal{L}_{t}$ is updated with newly acquired $10$ iD samples and the selected non-iD samples until acquiring the minimum number of iD samples. Therefore, the updated samples including all of iD, ambiguous, and OoD samples are removed from $\mathcal{P}^\mathcal{U}_{t}$. The annotation costs, i.e., the number of iD and non-iD samples examined by the annotator, are accumulated. For training the classification model, we undersample the labeled non-iD samples because of performance degradation caused by the data imbalance issue. The active learning process continues until the number of labeled iD samples in $\mathcal{P}^\mathcal{L}_t$ reaches to $300$ for the MixMNIST experiment.

For MixCIFAR60, the number of samples in the initial unlabeled data pool $\mathcal{P}^\mathcal{U}_0$ is $60,000$ as can be seen in Table~\ref{tab:dataset-stat}. The number of the initial labeled iD samples is set to $2,000$ and that of iD samples that should be acquired at a single stage is also set to $2,000$. The active learning process is terminated when the number of labeled iD samples in $\mathcal{P}^\mathcal{L}_t$ reaches to $20,000$.

\subsection{Results}
In this section, we provide the experimental results on the proposed active learning benchmarks.

\subsubsection{MixMNIST.} As shown in Figure~\ref{fig:perf-mixedmnist}, least confidence and entropy sampling select a lot of non-iD samples to acquire the predefined number of iD samples, especially at the early stage where the model has not learned non-ID samples yet. Although random sampling is not a suitable acquisition strategy in terms of accuracy, it can be considered to be efficient in terms of the annotation costs. Our acquisition strategy, Distance(CL), shows outstanding active learning performance, particularly in terms of the annotation costs. Table~\ref{tab:performance} provides the annotation costs required to increase accuracy by $1\%$. Least confidence sampling yields the best accuracy among the comparative methods, but it needs $68.00$ samples to increase accuracy by $1\%$ while entropy sampling requires $348.68$ samples on average. On the other hand, all contrastively learned models demand only less than $5$ samples. It is significantly lower than the comparative methods. Specifically, Distance(CL) consumes the lowest annotation costs, $406$ samples, among the comparative methods while showing comparable accuracy to least confidence sampling. It demonstrates that the proposed active learning method successfully acquires informative iD samples from the realistic data pool.

It is worth noting that, while updating $\mathcal{P}^\mathcal{L}_1$, we take into account all iD, ambiguous, and OoD samples of $\mathcal{P}^\mathcal{U}_0$ in the case of the contrastive learning approach, but only iD samples of $\mathcal{P}^\mathcal{U}_0$ are taken into account in the case of the comparative methods. 
Despite this disadvantage, the proposed method can effectively choose informative iD samples even at the early stage, resulting in much lower annotation costs.

\subsubsection{MixCIFAR60.} Similar to the results on MixMNIST, the comparative methods such as least confidence and entropy sampling are costly. DDU, which is devised to distinguish informative iD samples from ambiguous samples, consumes the lowest annotation costs among the comparative methods. The final accuracy values of contrastively learned models are slightly low, but they show better performance across all active learning stages when the same amount of annotation budgets are given as shown in Figure~\ref{fig:perf-mixedcifar60}.

Table~\ref{tab:performance} demonstrates the annotation cost efficiency of the contrastive learning approaches. To improve accuracy by $1\%$, Random(CL) requires $505.41$ samples and Distance(CL) consumes the lowest annotation costs, which is only $420.17$ samples. In addition, the contrastively learned models achieve high accuracy at early stages thanks to self-supervised representation learning. In other words, they can reach competitive accuracy even when we can access only a few labeled iD samples. Note that we did not search the optimal hyperparameters for contrastive learning although it is sensitive to those hyperparameters. We believe that hyperparameter searching can bring additional performance improvement.

\begin{table}
    \centering
    \caption{The number of samples to be labeled for improving accuracy by $1\%$. Acc. and Cost are the classification accuracy and the cumulative annotation costs at the last active learning stage, respectively. Acquisition Func. indicates the acquisition function. \textbf{Bold} represents the best performance among the comparative methods on each benchmark.}
    \label{tab:performance}
    \begin{tabular}{c|c|c c|c} 
    \toprule
    & \textbf{Acquisition Func.} & \textbf{Acc.} & \textbf{Cost} & \textbf{Cost/Acc.} \\
    \midrule
    \multirow{5}{*}{\rotatebox[origin=c]{90}{MixMNIST}} & Random & 93.84 & 440 & 4.69 \\
    & Least Confidence & \textbf{96.80} & 6,582 & 68.00 \\
    & Entropy & 93.89 & 32,738 & 348.68 \\
    \cmidrule{2-5}
    & {Random (CL)} & 96.07 & 420 & 4.37 \\
    & {Distance (CL)} & 96.65 & \textbf{406} & \textbf{4.20} \\
    
    \midrule
    \multirow{6}{*}{\rotatebox[origin=c]{90}{MixCIFAR60}} & Random & 67.95 & 39,270 & 577.92 \\
    & Least Confidence & 69.90 & 42,676 & 610.53 \\
    & Entropy & \textbf{70.43} & 42,678 & 605.96 \\
    & DDU & 67.38 & 32,162 & 477.32 \\
    \cmidrule{2-5}
    & {Random (CL)} & 68.38 & 34,560 & 505.41 \\
    & {Distance (CL)} & 66.38 & \textbf{27,891} & \textbf{420.17} \\
    
    \bottomrule
    \end{tabular}
\end{table}

\section{Conclusion}
In this paper, we argue the necessity of an active learning benchmark reflecting real-world data gathering processes and discuss the undesirable behavior of the commonly employed active learning methods in this situation. Therefore, we propose two active learning benchmarks, MixMNIST and MixCIFAR60, consisting of samples from various distributions. The proposed benchmarks are comprised of three categories: iD, ambiguous, and OoD samples. In addition, we propose a novel active learning method that works effectively on the realistic benchmark, which employs contrastive loss to leverage both labeled and unlabeled data pools. It is empirically validated that the contrastively learned feature space preserves the semantic relationship between samples from different distributions and classes. Our acquisition strategy selects informative iD samples from clusters based on the distance between features in the learned feature space. With the proposed benchmarks, the experimental results demonstrate the limitation of the commonly employed active learning methods and the effectiveness of the proposed approach in terms of annotation costs and task performance.

\subsubsection{Discussion.}
Although our proposed method significantly improves the classification performance for MixMNIST, it does not provide the same amount of improvement for MixCIFAR60. We infer that it is due to the distance threshold employed as the basis for our sample acquisition strategy. One of our assumptions is that the features of non-iD samples will be placed further to the cluster centroid than those of iD samples. However, if this assumption does not hold, i.e., one of the non-iD samples is located very close to the centroid, uninformative samples are acquired. Since MixCIFAR60 is a much more challenging benchmark than MixMNIST, the representation model may struggle to produce high-quality feature representations that preserve the semantic distances of samples belonging to each category. This motivates the investigation of an acquisition strategy that is robust to the quality of feature representation. We leave it for future work.


\bibliography{aaai23}






\end{document}